\documentclass[a4paper]{report}
\usepackage[utf8]{inputenc}
\usepackage[T1]{fontenc}
\usepackage{RJournal}
\usepackage{amsmath,amssymb,array}
\usepackage{booktabs}

\begin{document}

\sectionhead{Contributed research article}
\volume{XX}
\volnumber{YY}
\year{20ZZ}
\month{AAAA}

\begin{article}
\title{MDFS - MultiDimensional Feature Selection}
\author{by Radosław Piliszek, Krzysztof Mnich, Szymon Migacz, Paweł Tabaszewski, Andrzej Sułecki, Aneta Polewko-Klim and Witold Rudnicki}

\maketitle

\renewcommand\cite\citep

\abstract{
Identification of informative variables in an information system is often performed using simple
one-dimensional filtering procedures that discard information about interactions between variables.
Such approach may result in removing some relevant variables from consideration.
Here we present an R package \CRANpkg{MDFS} (MultiDimensional Feature Selection) that performs
identification of informative variables taking into account synergistic interactions between
multiple descriptors and the decision variable.
MDFS is an implementation of an algorithm based on information theory \cite{DBLP:journals/corr/MnichR17}.
Computational kernel of the package is implemented in C++.
A high-performance version implemented in CUDA C is also available.
The applications of \CRANpkg{MDFS} are demonstrated using the well-known Madelon dataset
that has synergistic variables by design.
The dataset comes from the UCI Machine Learning Repository \cite{Dua:2017}.
It is shown that multidimensional analysis is more sensitive than one-dimensional tests
and returns more reliable rankings of importance.
}

\section{Introduction}

Identification of variables that are related to the decision variable is often the most important step in dataset analysis.
In particular, it becomes really important when the number of variables describing the phenomena under scrutiny is large.

Methods of feature selection fall into three main categories \cite{guyon2003introduction}:
\begin{itemize}
\item filters, where the identification of informative variables is performed before data modelling and analysis,
\item wrappers, where the identification of informative variables is achieved by analysis of the models,
\item embedded methods, which evaluate utility of variables in the model and select the most useful variables.
\end{itemize}

Filters are designed to provide a quick answer and therefore are the fastest.
On the other hand, their simplicity is also the source of their errors.
The rigorous univariate methods, such as t-test, don't detect interactions between variables.
Heuristical methods that avoid this trap, such as Relief-f algorithm \cite{kononenko1994estimating}, may be biased towards weak and correlated variables \cite{robnik2003theoretical}. Several filtering methods are designed to return only the non-redundant subset of variables \cite{zhao2007searching,peng2005feature,wang2013selecting}.
While such methods may lead to very efficient models, their selection may be far from the best when one is interested in deeper understanding of the phenomena under scrutiny.

The wrapper algorithms are designed around machine learning algorithms such as SVM \cite{cortes1995support}, as in the SVM-RFE algorithm \cite{guyon2002gene}, or random forest \cite{Breiman2001}, as in the Boruta algorithm \cite{Kursa2010a}.
They can identify variables involved in non-linear interactions.
Unfortunately, for systems with tens of thousands of variables they are slow.
For example, the Boruta algorithm first expands the system with randomised copies of variables and then requires numerous runs of the random forest algorithm.

The embedded methods are mostly limited to linear approximations and are part of a modelling approach where the selection is directed towards the utility of the model \cite{lasso,elastic_net}.
Therefore, variables that are relevant for understanding the phenomena under scrutiny may be omitted and replaced by variables more suitable for building a particular model.

\section{Theory}

Kohavi and John proposed that a variable $x_i \in X$, where $X$ is a set
of all descriptive variables, is weakly relevant if there exists such
subset of variables $X_{sub} \subset X : x_i \notin X_{sub}$
that one can increase information on the decision variable $y$ by
extending this subset with the variable $x_i$ \cite{kohavi_john}.
Mnich and Rudnicki introduced the notion of $k$-weak relevance,
that restricts the original definition by Kohavi and John to
$(k-1)$-element subsets $X_{sub}$ \cite{DBLP:journals/corr/MnichR17}.

The algorithm implements the definition of $k$-weak relevance directly
by exploring all possible $k$-tuples of variables
$x_i~\cup~\{x_{m_1},x_{m_2},\ldots,x_{m_{k-1}}\}$
for $k$-dimensional analysis.
The maximum decrease in conditional information entropy upon adding
$x_i$ to description, normalized to sample size, is used as the measure
of $x_i$'s relevance:
\begin{equation}\label{eq:IGmax}
IG^k_{max}(y;x_i) = N \max_m\left(H(y|x_{m_1},x_{m_2},\ldots,x_{m_{k-1}}) - H(y|x_i,x_{m_1},x_{m_2},\ldots,x_{m_{k-1}})\right),
\end{equation}
where $H$ is (conditional) information entropy and $N$ is the number of observations.
Difference in (conditional) information entropy is known as (conditional) mutual information.
It is multiplied by $N$ to obtain the proper null-hypothesis distribution.
To name this value we reused the term information gain ($IG$) which is
commonly used in information-theoretic context to denote different values
related to mutual information.

To declare a variable $k$-weakly relevant it is required that its
$IG^k_{max}(y;x_i)$ is statistically significant.
This can be established via a comparison:
\begin{equation}
IG^k_{max}(y;x_i) \geq IG_{lim},
\end{equation}
where $IG_{lim}$ is computed using a procedure of fitting the
theoretical distribution to the data.

For a sufficiently large sample, the value of $IG$ for a non-informative variable,
with respect to a single $k$-tuple, follows a $\chi^2$ distribution.
$IG^k_{max}(y;x_i)$, which is the maximum value of $IG$ among many trials, follows an extreme value distribution.
This distribution has one free parameter corresponding to the number of independent tests
which is generally unknown and smaller than the total number of tests.
The parameter is thus computed empirically by fitting the distribution
to the irrelevant part of the data \cite{DBLP:journals/corr/MnichR17}.
This allows to convert the $IG^k_{max}$ statistic to its $p$-value and then to establish $IG_{lim}$
as a function of significance level $\alpha$.
Since many variables are investigated, the $p$-value should be adjusted using well-known
FWER \cite{holm} or FDR \cite{benjamini_hochberg} control technique.
Due to unknown dependencies between tests, for best results we recommend using Benjamini-Hochberg-Yekutieli
method \cite{benjamini2001}\footnote{Method \code{"BY"} for \code{p.adjust} function.} when performing FDR.

In one dimension ($k = 1$) Eq.~\ref{eq:IGmax} reduces to:
\begin{equation}
IG^1_{max}(y;x_i) = N(H(y) - H(y|x_i)),
\end{equation}
which is a well-known G-test statistic \cite{sokal94}.

All variables that are weakly relevant in one-dimensional test should also be discovered in higher-dimensional tests,
nevertheless their relative importance may be significantly influenced by interactions with other variables.
Often the criterium for inclusion to further steps of data analysis and model building is simply taking top $n$ variables,
therefore the ordering of variables due to importance matters as well.

\section{Algorithm and implementation}

The MDFS package consists of two main parts.
The first one is an R \cite{RManual} interface to two computational engines.
These engines utilise either CPU or NVIDIA GPU and are implemented in standard C++ and in CUDA C, respectively.
Either computational engine returns the $IG^k_{max}$ distribution for a given dataset plus requested details which may pose an interesting insight into data.
The second part is a toolkit to analyse results.
It is written entirely in R.
The version of the \CRANpkg{MDFS} package used and described here is 1.0.2.

The $IG^k_{max}$ for each variable is computed using a straightforward algorithm based on Eq.~\ref{eq:IGmax}.
Information entropy ($H$) is computed using discretised descriptive variables.
Discretisation is performed using customisable randomised rank-based approach.
To control the discretisation process we use a concept of range.
Range is a real number between 0 and 1 affecting the share each discretised variable class has in the dataset.
Each share is sampled from a uniform distribution on the interval $(1 - \mbox{range}, 1 + \mbox{range})$.
Hence, $\mbox{range} = 0$ results in an equipotent split, $\mbox{range} = 1$ equals a completely random split.
Let's assume that there are $N$ objects in the system and we want to discretise a variable to $k$ classes.
To this end, $(k-1)$ distinct integers from the range \((2,N)\) are obtained using computed shares.
Then, the variable is sorted and values at positions indexed by these integers are used to discretise the variable into separate classes.
In most applications of the algorithm there is no default best discretisation of descriptive variables,
hence multiple random discretisations are performed.
The $IG^k_{max}$ is computed for each discretisation, then the maximum information gain obtained in any
of the discretisations is returned. Hence, the $IG^k_{max}$ is maximum over tuples and discretisations.

Conditional information entropy is obtained from the experimental probabilities of decision class using the following formula:
\begin{equation}
H(y|x_1,\ldots,x_k) = -\sum_{d=0,1} \sum_{i_1=1:c} \ldots \sum_{i_k=1:c} p^{d}_{i_1,\ldots,i_k}\log\left(p^{d}_{i_1,\ldots,i_k}\right)
\end{equation}
where $p^{d}_{i_1,\ldots,i_k}$ denotes the conditional probability of class $d$ in a \mbox{$k$-dimensional} voxel with coordinates $i_j$.
To this end, one needs to compute the number of instances of each class in each voxel.
The conditional probability of class $d$ in a voxel is then computed as
\begin{equation}
p^{d}_{i_1,\ldots,i_k} = \frac{N^d_{i_1,\ldots,i_k}+\beta^k}{N^0_{i_1,\ldots,i_k}+\beta^0+N^1_{i_1,\ldots,i_k}+\beta^1},
\end{equation}
where $N^d_{i_1,\ldots,i_k}$ is the count of class $d$ in a \mbox{$k$-dimensional voxel} with coordinates $i_j$
and $\beta^d$ is pseudocount corresponding to class $d$:
\begin{equation}
\beta^d = \xi \frac{N^d}{\min(N^0,N^1)}
\end{equation}
where $\xi \in (0,1)$ is supplied by the user.
Note that the number of voxels in $k$ dimensions is $N=c^k$, where $c$ is the number of classes of discretised descriptive variables.

The implementation of the algorithm is currently limited to binary decision variables.
The analysis for information systems that have more than two categories can be
performed either by executing all possible pairwise comparisons or one-vs-rest.
Then all variables that are relevant in the context of a single pairwise comparison should be considered relevant.
In the case of continuous decision variable one must discretise it before performing analysis.
In the current implementation all variables are discretised into an equal number of classes.
This constraint is introduced for increased efficiency of computations, in particular on GPU.

Another limitation is the maximum number of dimensions set to 5.
This is due to several reasons.
Firstly, the computational cost of the algorithm is proportional to number
of variables to power equal the dimension of the analysis, and it becomes
prohibitively expensive for powers larger than 5 even for systems described with
a hundred of variables.
Secondly, analysis in higher dimensions requires a substantial number of objects
to fill the voxels sufficiently for the algorithm to detect real synergies.
Finally, it is also related to the simplicity of efficient implementation of
the algorithm in CUDA.
The most time consuming part of the algorithm is computing the counters for all voxels.
Fortunately, this part of computations is relatively easy to parallelise, as the
exhaustive search is very well suited for GPU.
Therefore, a GPU version of the algorithm was developed in CUDA C for NVIDIA GPGPUs
and is targeted towards problems with a very large number of features.
The CPU version is also parallelised to utilise all cores available on a single node.
The 1D analysis is available only in the CPU version since there is no benefit
in running this kind of analysis on GPU.

\section{Examples}

\subsection{Package functions introduction}

There are three functions in the package which are to be run directly with the input dataset: \code{MDFS}, \code{ComputeMaxInfoGains} and \code{ComputeInterestingTuples}.
The first one, \code{MDFS}, is our recommended function for new users since it hides
internal details and provides an easy to use interface for basic end-to-end analysis
for current users of other statistical tests (e.g. \code{t.test}) so that the user
can straightforwardly get the statistic values, p-values and adjusted p-values
for variables from input.
The other two functions are interfaces to the IG-calculating lower-level C++ and CUDA C++ code.
\code{ComputeMaxInfoGains}  returns the max IGs as described in the theory section.
It can optionally provide information about the tuple in which this max IG was observed.
On the other hand, one might be interested in tuples where certain IG threshold has been achieved.
The \code{ComputeInterestingTuples} function performs this type of analysis and reports
which variable in which tuple achieved the corresponding IG value.

The \code{ComputePValue} function performs fitting of IGs to respective statistical distributions
as described in the theory section and returns object of the \code{MDFS} class including,
in particular, p-values for variables.
This class implements various methods for handling output of statistical analysis.
In particular they can plot details of IG distribution, output p-values of all variables, output relevant variables.
\code{ComputePValue} is implemented in a very general way, extending beyond limitations of the current implementation of \code{ComputeMaxInfoGains}.
In particular, it can handle multi-class problems and different number of divisions for each variable.

The \code{AddContrastVariables} is an utility function used to construct contrast variables \cite{Stoppiglia2003,Kursa2010a}.
Contrast variables are used for improving reliability of the fit of statistical distribution.
In the case of fitting distribution to contrast variables we know exactly how many irrelevant variables there are in the system.
The contrast variables are not taken into account when computing adjusted p-values to avoid decreasing the sensitivity.

\subsection{Canonical package usage}

As mentioned earlier, the recommended way to use the package is to use the \code{MDFS} function.
It uses the other packaged functions to achieve its goal in the standard and thoroughly tested way,
so it may be considered the canonical package usage pattern.
Hence, let us examine the code of the \code{MDFS} function in a step-by-step manner.

The first line sets the passed seed so that it is used both during the contrast variables construction and computing IGs:
\begin{example}
if (!is.null(seed)) set.seed(seed)
\end{example}
The seed used for IG calculation is set and preserved in attributes of \code{MIG.Result} for reproducibility of results.
The information about which variables were used to contrast variables is also preserved.

In the next step the function actually builds the contrast variables (if not disabled)
and sets the whole feature set (\code{data.contrast}) to be used in further functions:
\begin{example}
if (n.contrast > 0) {
  contrast <- AddContrastVariables(data, n.contrast)
  contrast.indices <- contrast$indices
  contrast.variables <- contrast$x[,contrast$mask]
  data.contrast <- contrast$x
  contrast.mask <- contrast$mask
} else {
  contrast.mask <- contrast.indices <- contrast.variables <- NULL
  data.contrast <- data
}
\end{example}

In the next step the compute-intensive computation of IGs is executed:
\begin{example}
MIG.Result <- ComputeMaxInfoGains(data.contrast, decision,
  dimensions = dimensions, divisions = divisions,
  discretizations = discretizations, range = range, pseudo.count = pseudo.count,
  seed = seed, return.tuples = !use.CUDA && dimensions > 1, use.CUDA = use.CUDA)
\end{example}
The first two positional parameters are respectively the feature data and the decision.
The other parameters decide on the type of computed IGs:
\code{dimensions} controls dimensionality,
\code{divisions} controls the number of classes in the discretisation (it is equal to \code{divisions+1}),
\code{discretizations} controls the number of discretisations,
\code{range} controls how random the discretisation splits are and
\code{pseudo.count} controls the regularization parameter (pseudocounts).

Finally, the computed IGs are analysed and a statistical result is computed and returned:
\begin{example}
divisions <- attr(MIG.Result, "run.params")$divisions

fs <- ComputePValue(MIG.Result$IG,
  dimensions = dimensions, divisions = divisions,
  contrast.mask = contrast.mask,
  one.dim.mode = ifelse (discretizations==1, "raw",
                         ifelse(divisions*discretizations<12, "lin", "exp")))

statistic <- if(is.null(contrast.mask)) { MIG.Result$IG }
             else { MIG.Result$IG[!contrast.mask] }
p.value <- if(is.null(contrast.mask)) { fs$p.value }
           else { fs$p.value[!contrast.mask] }
adjusted.p.value <- p.adjust(p.value, method = p.adjust.method)
relevant.variables <- which(adjusted.p.value < level)
\end{example}
In the first line \code{divisions} is set from attributes of \code{MIG.Result} because it is adjusted by the algorithm when left unset by the user.
The \code{one.dim.mode} parameter controls the expected distribution in 1D.
The rule states that as long as we have 1 discretisation the resulting distribution is chi-squared,
otherwise, depending on the product of \code{discretizations} and \code{divisions}, the resulting
distribution might be closer to a linear or exponential, as in higher dimensions, function of chi-squared distributions.
This is heuristic and might need to be tuned.
Features with adjusted p-values below some set level are considered to be relevant.

\subsection{Madelon example}

For demonstration of the MDFS package we used the training subset of the well-known Madelon dataset \cite{guyon2007competitive}.
It is an artificial set with 2000 objects and 500 variables.
The decision was generated using a 5-dimensional random parity function based on variables drawn from normal distribution.
The remaining variables were generated in the following way.
Fifteen variables were obtained as linear combinations of the 5 input variables and remaining 480 variables were drawn randomly from the normal distribution.
The data set can be accessed from the UCI Machine Learning Repository \cite{Dua:2017} and it is included in \CRANpkg{MDFS} package as well.

We conducted the analysis in all possible dimensionalities using both CPU and GPU versions of the code.
Additionally, a standard t-test was performed for reference.
We examined computational efficiency of the algorithm and compared the results obtained by performing analysis in varied dimensionalities.

In the first attempt we utilised the given information on the properties of the dataset under scrutiny.
We knew in advance that Madelon was constructed as a random parity problem and that each base variable was constructed from a distinct distribution.
Therefore, we could use one discretisation into 2 equipotent classes.
In the second attempt the recommended 'blind' approach in 2D was followed which utilises several randomized discretisations.

For brevity, in the following examples the set of Madelon independent variables is named \code{x} and its decision is named \code{y}:
\begin{example}
x <- madelon$data
y <- madelon$decision
\end{example}

For comparison of our approach with the t-test a simple wrapper to get p-values for all features is introduced:
\begin{example}
t.test.all <- function(x, y) {
  do.t.test <- function(i, x1, x2) {
    return(t.test(x1[,i], x2[,i])$p.value)
  }
  sapply(1:ncol(x), do.t.test, x[y == T,], x[y == F,])
}
\end{example}

One can now obtain p-values from t-test, adjust them using
Holm correction (one of FWER corrections, the default in the \code{p.adjust} function),
take relevant with level $0.05$ and order them:
\begin{example}
> tt <- t.test.all(x, y)
> tt.adjusted <- p.adjust(tt, method = "holm")
> tt.relevant <- which(tt.adjusted < 0.05)
> tt.relevant.ordered <- tt.relevant[order(tt.adjusted[tt.relevant])]
> tt.relevant.ordered
[1] 476 242 337  65 129 106 339  49 379 443 473 454 494
\end{example}
A FWER correction is used because we expect strong separation between relevant and irrelevant features in this artificial dataset.

To achieve the same with \CRANpkg{MDFS} for 1, 2 and 3 dimensions one can use the wrapper \code{MDFS} function:
\begin{example}
> d1 <- MDFS(x, y, n.contrast = 0, dimensions = 1, divisions = 1, range = 0)
> d1.relevant.ordered <- d1$relevant.variables[order(d1$p.value[d1$relevant.variables])]
> d1.relevant.ordered
[1] 476 242 339 337  65 129 106  49 379 454 494 443 473
> d2 <- MDFS(x, y, n.contrast = 0, dimensions = 2, divisions = 1, range = 0)
> d2.relevant.ordered <- d2$relevant.variables[order(d2$p.value[d2$relevant.variables])]
> d2.relevant.ordered
[1] 476 242  49 379 154 282 434 339 494 454 452  29 319 443 129 473 106 337  65
> d3 <- MDFS(x, y, n.contrast = 0, dimensions = 3, divisions = 1, range = 0)
> d3.relevant.ordered <- d3$relevant.variables[order(d3$p.value[d3$relevant.variables])]
> d3.relevant.ordered
[1] 154 434 282  49 379 476 242 319  29 452 494 106 454 129 473 443 339 337  65 456
\end{example}
The changes in the relevant variables set can be examined with simple \code{setdiff} comparisons:
\begin{example}
> setdiff(tt.relevant.ordered, d1.relevant.ordered)
integer(0)
> setdiff(d1.relevant.ordered, tt.relevant.ordered)
integer(0)
> setdiff(d1.relevant.ordered, d2.relevant.ordered)
integer(0)
> setdiff(d2.relevant.ordered, d1.relevant.ordered)
[1] 154 282 434 452  29 319
> setdiff(d2.relevant.ordered, d3.relevant.ordered)
integer(0)
> setdiff(d3.relevant.ordered, d2.relevant.ordered)
[1] 456
\end{example}
One may notice that ordering by importance leads to different results for these 4 tests.

In the above the knowledge about properties of the Madelon dataset was used: that there are many random variables,
hence no need to add contrast variables, and that the problem is best resolved by splitting features in half,
hence one could use 1 discretisation and set range to zero.

However, one is usually not equipped with such knowledge and then may need to use multiple random discretisations.
Below an example run of 'blinded' 2D analysis of Madelon is presented:
\begin{example}
> d2b <- MDFS(x, y, dimensions = 2, divisions = 1, discretizations = 30, seed = 118912)
> d2b.relevant.ordered <- d2b$relevant.variables[order(d2b$p.value[d2b$relevant.variables])]
> d2b.relevant.ordered
[1] 476 242 379  49 154 434 106 282 473 339 443 452  29 454 494 319  65 337 129
> setdiff(d2b.relevant.ordered, d2.relevant.ordered)
integer(0)
> setdiff(d2.relevant.ordered, d2b.relevant.ordered)
integer(0)
\end{example}
This demonstrates that the same variables are discovered, yet with a different order.

\section{Results}

\subsection{Performance}

The performance of the CPU version of the algorithm was measured on a computer
with two Intel Xeon E5-2650 v2 processors, running at 2.6 GHz.
Each processor has eight physical cores. Hyperthreading was disabled.

The GPU version was tested on a computer with a single NVIDIA Tesla K80 accelerator.
The K80 is equipped with two GK210 chips and is therefore visible to the system as two separate GPGPUs.
Both were utilised in the tests.

The Madelon dataset has moderate dimensionality for modern standards, hence it is amenable to high-dimensional analysis.
The CPU version of the code handles analysis up to four dimensions in a reasonable time, see Table~\ref{Madelon:times}.

The performance gap between CPU and GPU versions is much higher than suggested
by a simple comparison of hardware capabilities.
This is due to two factors.
Firstly, the GPU version has been highly optimised towards increased efficiency of memory usage.
The representation of the data by bit-vectors and direct exploitation of the data locality allows for much higher data throughput.
What is more, the bit-vector representation allows for using very efficient \emph{popcnt} instruction for counter updates.
On the other hand the CPU version has been written mainly as a reference version using a straightforward implementation of the algorithm and has not been strongly optimised.

\begin{table}[h]
\caption{Execution times for the Madelon dataset.}
\begin{center}
\begin{tabular}{l| r r r r r r }
\hline
		 & t-test  & 1D      & 2D     & 3D    & 4D    & 5D
\\\hline
 CPU & 0.21s   & 0.01s   & 0.44s  & 42s   & 2h    & 249h
\\
 GPU & -       & -       & 0.23s  & 0.2s  & 9.8s  & 1h
\\
\hline
\end{tabular}
\end{center}
\label{Madelon:times}
\end{table}

\begin{table}[h]
\caption{Execution times for the Madelon dataset with 30 random discretisations.}
\begin{center}
\begin{tabular}{l| r r r r r }
\hline
			& 1D     & 2D     & 3D    & 4D       & 5D
\\\hline
	CPU & 0.35s  & 5.8s   & 37min & 92h      & -
\\
	GPU & -      & 2.9s   & 3.3s  & 7min 36s & 42h
\\
\hline
\end{tabular}
\end{center}
\label{Madelon:times30}
\end{table}

\subsection{Structure of Madelon set revealed by MDFS analysis}

\begin{figure}[h]
\centering
\includegraphics[width=\linewidth]{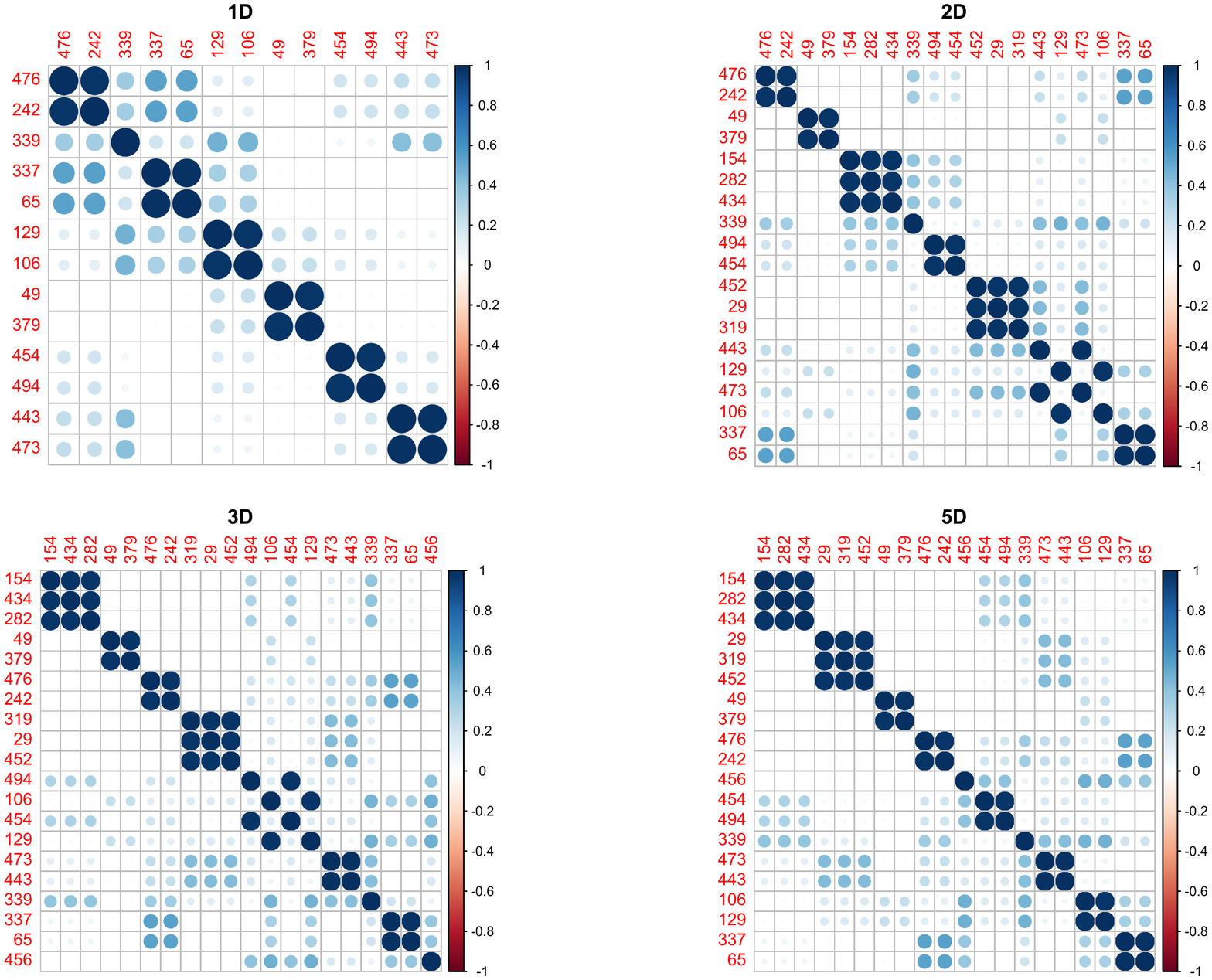}
\caption{
Correlation plots for relevant variables discovered in 1-, 2-, 3- and 5-dimensional analysis of the Madelon dataset with one deterministic discretisation with division in the middle.
The variables are ordered by IG.
}
\label{results:madelon1}
\end{figure}

\begin{table}[h!]
	\caption{Discovered variable clusters (as seen in correlation plots) ordered by descending maximum relevance (measured with 5D IG), identified by the variable with the lowest number.}
	\begin{center}
	\begin{tabular}{l| l }
	\hline
	Cluster & Members
	\\\hline
	154 & 154, 282, 434
	\\\hline
	29  & 29, 319, 452
	\\\hline
  49  & 49, 379
	\\\hline
	242 & 476, 242
	\\\hline
	456 & 456
	\\\hline
	454 & 454, 494
	\\\hline
  339 & 339
	\\\hline
  443 & 473, 443
	\\\hline
  106 & 106, 129
	\\\hline
  65  & 337, 65
	\\\hline
	\end{tabular}
	\end{center}
	\label{results:clusters}
\end{table}

\begin{figure}[h]
\centering
\includegraphics[trim={ 0.1cm  0.2cm 0.2cm 0.0cm},clip,width=0.95\linewidth]{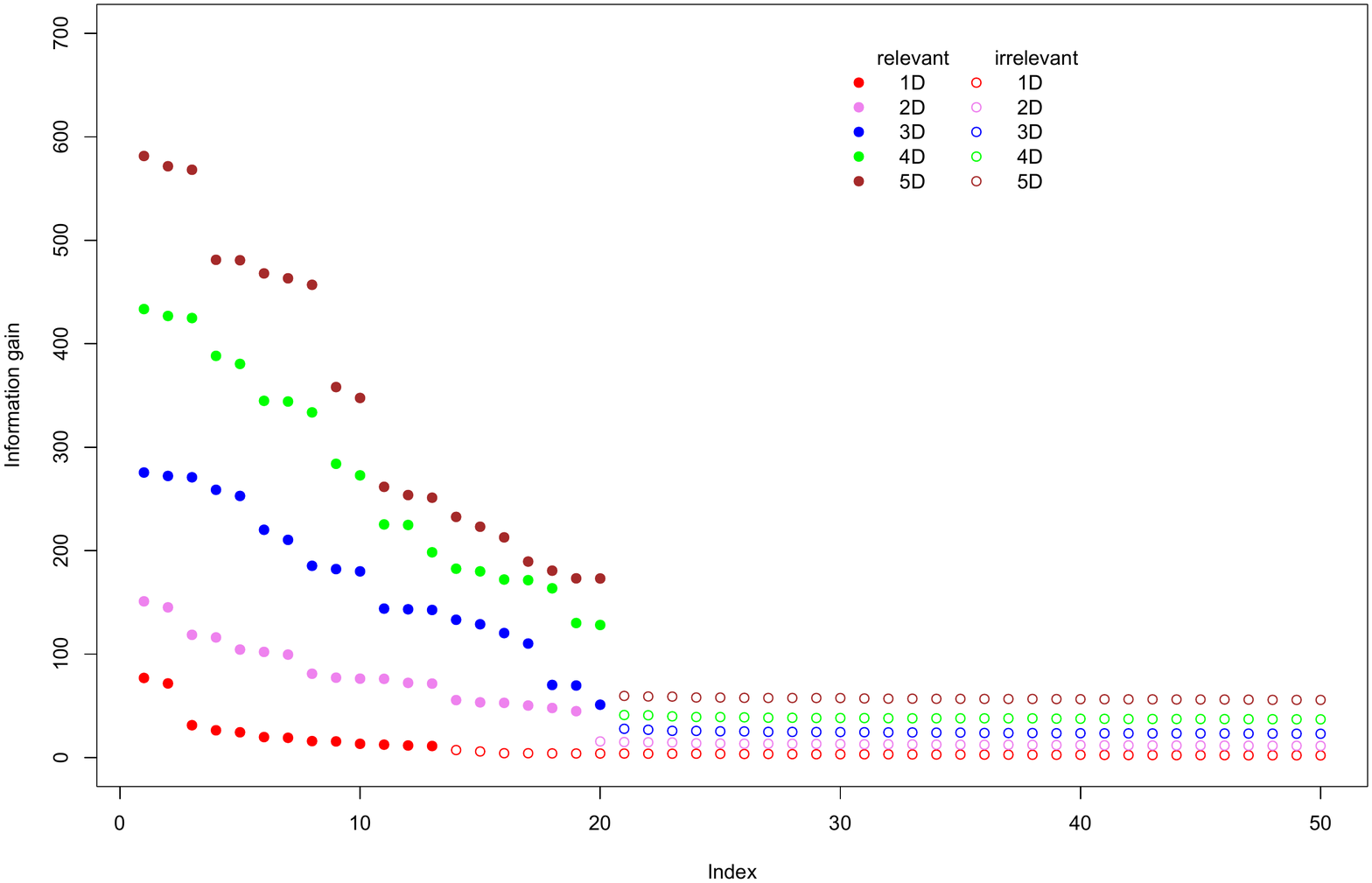}
\caption{
Information gain obtained by the MDFS algorithm using 1-, 2-, 3-, 4- and 5-dimensional variants
of the algorithm for the Madelon dataset with one deterministic discretisation with division in the middle.
Full circles represent variables deemed relevant.
All variables are sorted by IG.
Margin between irrelevant and relevant features grows with dimensionality.
}
\label{results:IG50}
\end{figure}

\begin{figure}[h]
\centering
\includegraphics[trim={ 0.1cm  0.2cm 0.2cm 0.0cm},clip,width=0.95\linewidth]{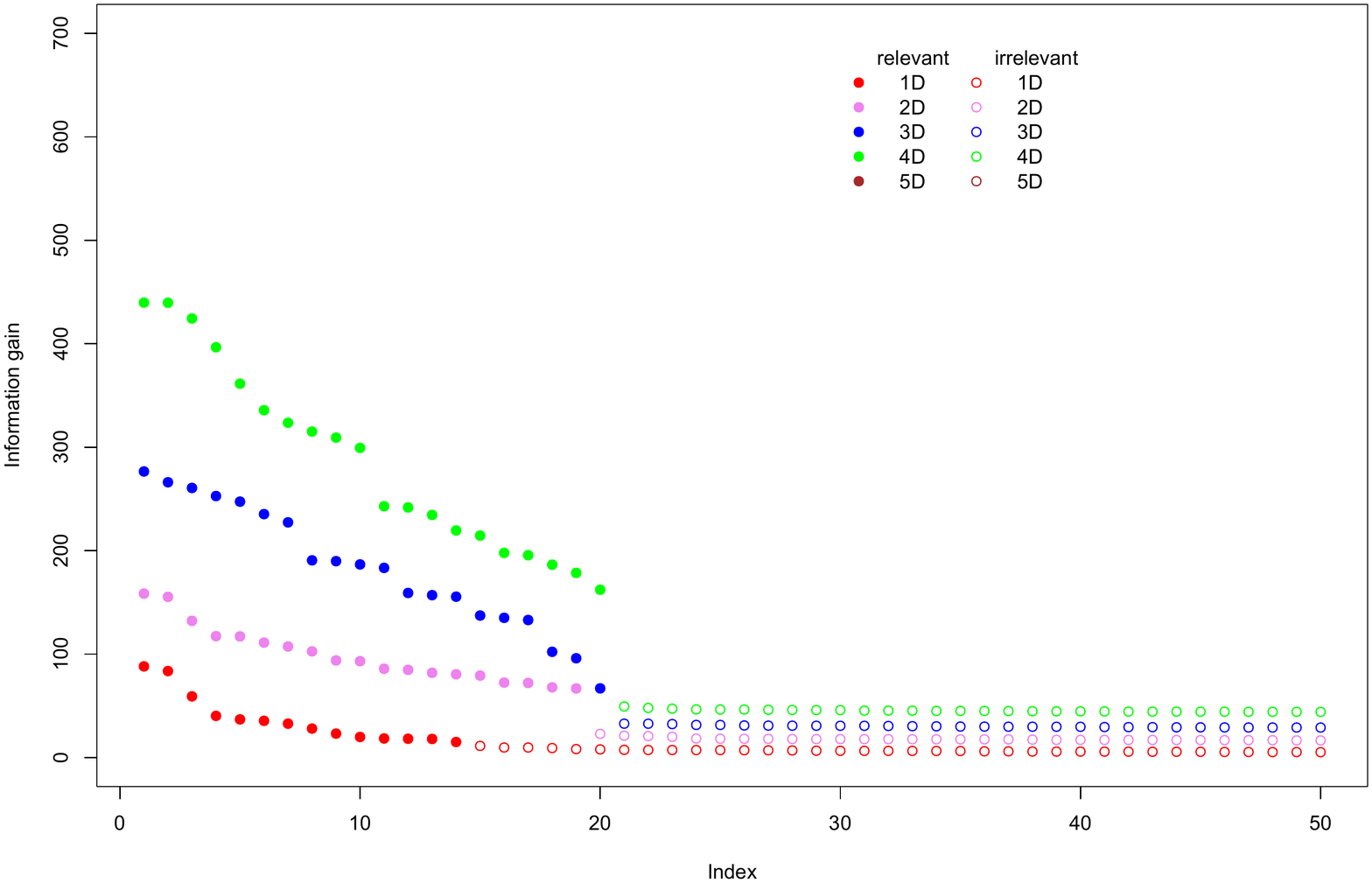}
\caption{
Information gain obtained by the MDFS algorithm using 1-, 2-, 3-, 4- and 5-dimensional variants
of the algorithm for the Madelon dataset with 30 random discretisations.
Full circles represent variables deemed relevant.
All variables are sorted by IG.
Margin between irrelevant and relevant features grows with dimensionality.
}
\label{results:IG50_blinded}
\end{figure}

\begin{table}[h!]
\caption{
Summary of results for the Madelon dataset with one deterministic discretisation with division in the middle.
The variable clusters are ordered by descending IG.
The numbers of base variable clusters are highlighted in boldface.
Clusters represented by 65 and 106, displayed in italic font, are deemed
highly relevant in 1D analyses and the least relevant in 5D analysis.
}
\begin{center}
\begin{tabular}{l| p{2.5em} p{2em} p{2em} p{2em} p{2em} p{2em} p{2em} }
\hline
		& t-test     & 1D        & 2D        & 3D        & 4D        & 5D
\\\hline
1.  & {\bf 242}  & {\bf 242} & {\bf 242} & {\bf 154} & {\bf 154} & {\bf 154}
\\\hline
2.  & {\it 65}   & 339       & {\bf 49}  & {\bf 49}  & {\bf 49}  & {\bf 29}
\\\hline
3.  & {\it 106}  & {\it 65}  & {\bf 154} & {\bf 242} & {\bf 29}  & {\bf 49}
\\\hline
4.  & 339        & {\it 106} & 339       & {\bf 29}  & {\bf 242} & {\bf 242}
\\\hline
5.  & {\bf 49}   & {\bf 49}  & 454       & 454       & 454       & {\bf 456}
\\\hline
6.  & 443        & 454       & {\bf 29}  & {\it 106} & 339       & 454
\\\hline
7.  & 454        & 443       & 443       & 443       & {\it 106} & 339
\\\hline
8.  & -          & -         & {\it 106} & 339       & {\bf 456} & 443
\\\hline
9.  & -          & -         & {\it 65}  & {\it 65}  & 443       & {\it 106}
\\\hline
10. & -          & -         & -         & {\bf 456} & {\it 65}  & {\it 65}
\\\hline
\end{tabular}
\end{center}
\label{Madelon:variables}
\end{table}

\begin{table}[h!]
	\caption{
	Summary of results for the Madelon dataset with 30 random discretisations.
	The variable clusters are ordered by descending IG.
	The numbers of base variable clusters are highlighted in boldface.
	Similar behaviour with 65 and 106 is observed as in the single discretisation case.
	Note the irrelevant variable 205 (underlined) discovered in 1D as relevant due to small margin between relevant and irrelevant features.
	}
	\begin{center}
	\begin{tabular}{l| p{2.5em} p{2em} p{2em} p{2em} p{2em} p{2em} p{2em} p{2em} }
	\hline
			& t-test    & 1D        & 2D        & 3D        & 4D        & 5D
	\\\hline
	1.  & {\bf 242} & {\bf 242} & {\bf 242} & {\bf 154} & {\bf 154} &
	\\\hline
	2.  & {\it 65}  & 339       & {\bf 49}  & {\bf 49}  & {\bf 49}  &
	\\\hline
	3.  & {\it 106} & {\it 65}  & {\bf 154} & {\bf 242} & {\bf 29}  &
	\\\hline
	4.  & 339       & 443       & {\it 106} & {\bf 29}  & {\bf 242} &
	\\\hline
	5.  & {\bf 49}  & {\it 106} & 443       & {\it 106} & 454       &
	\\\hline
	6.  & 443       & 454       & 339       & 454       & {\it 106} &
	\\\hline
	7.  & 454       & {\bf 49}  & {\bf 29}  & 443       & 339       &
	\\\hline
	8.  & -         & \underline{205} & 454       & 339       & 443       &
	\\\hline
	9.  & -         & -         & {\it 65}  & {\it 65}  & {\bf 456} &
	\\\hline
	10. & -         & -         & -         & {\bf 456} & {\it 65}  &
	\\\hline
	\end{tabular}
	\end{center}
	\label{Madelon:variables30}
\end{table}

The twenty relevant variables in Madelon can be easily identified by analysis of histograms
of variables, their correlation structure and by a priori knowledge of the method of construction of the dataset.
In particular, base variables, i.e. these variables that are directly connected to a decision variable,
have the unique distribution that has two distinct peaks.
All other variables have smooth unimodal distribution, hence identification of base variables is trivial.
What is more, we know that remaining informative variables are constructed as linear combinations of base variables,
hence they should display significant correlations with base variables.
Finally, the nuisance variables are completely random, hence they should not be correlated neither with base variables
nor with their linear combinations.
The analysis of correlations between variables reveals also that there are several groups of very highly correlated
($r>0.99$) variables, see Figure~\ref{results:madelon1}.
All variables in such a group can be considered as a single variable, reducing the number of independent variables to ten.
The entire group is further represented by the variable with the lowest number.
The clusters are presented in Table~\ref{results:clusters}.

This clear structure of the dataset creates an opportunity to confront results of the MDFS analysis
with the ground truth and observe how the increasing precision of the analysis helps to discover
this structure without using the a priori knowledge on the structure of the dataset.

One-dimensional analysis reveals 13 really relevant variables (7 independent ones),
both by means of the t-test and using the information gain measure, see Table~\ref{Madelon:variables}.
Three-dimensional and higher-dimensional analyses find all 20 relevant variables.
Additionally, with the exception of one-dimensional case, in all cases there is a clear separation
between IG obtained for relevant and irrelevant variables, see Figure~\ref{results:IG50}.
This translates into a significant drop of p-value for the relevant variables.

Five variables, namely $\{29,49,154,242,456\}$ are clearly orthogonal to each other,
hence they are the base variables used to generate the decision variable.
Five other variables are correlated with base variables and with each other,
and hence they are linear combinations of base variables.
The one-dimensional analyses, both t-test and mutual information approach,
find only two base variables, see Table~\ref{Madelon:variables}.
What is more, while one of them is regarded as highly important (lowest p-value)
- the second one is considered only the 5th most important out of 7.
Two-dimensional analysis finds 4 or 5 base variables, depending on the method used.
Moreover, the relative ranking of variables is closer to intuition, with three base variables on top.
The relative ranking of importance improves with increasing dimensionality of the analysis.
In 5-dimensional analysis all five base variables are scored higher than any linear combination.
In particular, the variable 456, which is identified by 3D analysis as the least important,
rises to the eight place in 4D analysis and to the fifth in 5D.
Interestingly, the variable 65, which is the least important in 5D analysis is the second
most important variable in t-test and the third most important variable in 1D.

\section{Conclusion}

We have introduced a new software library for identification of informative variables in multidimensional information systems which takes into account interactions between variables.
The implemented method is significantly more sensitive than the standard t-test when interactions between variables are present in the system.
When applied to the well-known five-dimensional problem Madelon the method not only discovered all relevant variables but also produced the correct estimate of their relative relevance.

\section*{Acknowledgments}

The research was partially funded by the Polish National Science Centre, grant 2013/09/B/ST6/01550.

\bibliography{piliszek-et-al}

\address{Radosław Piliszek\\
  Computational Centre, University of Bialystok\\
  Konstantego Ciolkowskiego 1M, 15-245 Bialystok\\
  Poland\\
  0000-0003-0729-9167\\
  \email{r.piliszek@uwb.edu.pl}}

\address{Krzysztof Mnich\\
  Computational Centre, University of Bialystok\\
  Konstantego Ciolkowskiego 1M, 15-245 Bialystok\\
  Poland\\
  0000-0002-6226-981X\\
  \email{k.mnich@uwb.edu.pl}}

\address{Szymon Migacz\\
  Interdisciplinary Centre for Mathematical and Computational Modelling, University of Warsaw\\
  Pawińskiego 5A, 02-106 Warsaw\\
  Poland}

\address{Paweł Tabaszewski\\
  Interdisciplinary Centre for Mathematical and Computational Modelling, University of Warsaw\\
  Pawińskiego 5A, 02-106 Warsaw\\
  Poland}

\address{Andrzej Sułecki\\
  Interdisciplinary Centre for Mathematical and Computational Modelling, University of Warsaw\\
  Pawińskiego 5A, 02-106 Warsaw\\
  Poland}

\address{Aneta Polewko-Klim\\
  Institute of Informatics, University of Bialystok\\
  Konstantego Ciolkowskiego 1M, 15-245 Bialystok\\
  Poland\\
  0000-0003-1987-7374\\
  \email{anetapol@uwb.edu.pl}}

\address{Witold Rudnicki\\
  Institute of Informatics, University of Bialystok\\
  Konstantego Ciolkowskiego 1M, 15-245 Bialystok\\
  Poland\\
  and\\
  Interdisciplinary Centre for Mathematical and Computational Modelling, University of Warsaw\\
  Pawińskiego 5A, 02-106 Warsaw\\
  Poland\\
  0000-0002-7928-4944\\
  \email{w.rudnicki@uwb.edu.pl}}

\end{article}

\end{document}